# Predicting Playa Inundation Using a Long Short-Term Memory Neural Network

Kylen Solvik[1,2,3], Anne M. Bartuszevige[4], Meghan Bogaerts[4], and Maxwell B. Joseph[1,2]

[1]North Central Climate Adaptation Science Center, Boulder, CO, USA

[2]Earth Lab, Cooperative Institute for Research in Environmental Sciences (CIRES), University of Colorado, Boulder, CO, USA

[3]Department of Geography, University of Colorado, Boulder, CO, USA

[4]Playa Lakes Joint Venture, Lafayette, CO, USA

Corresponding author: Kylen Solvik (kylen.solvik@coloardo.edu)

**Key Points:**

- Playas are infrequently-inundated critical wetland habitats for migratory birds and a source of recharge for the High Plains aquifer.
- Modeling playa inundation is challenging but highly valuable for conservation efforts, especially under a changing climate.
- Using a convolutional neural network, we can accurately predict monthly inundation for 71,842 playas across the Great Plains.

**Abstract**

In the Great Plains, playas are critical wetland habitats for migratory birds and a source of recharge for the agriculturally-important High Plains aquifer. The temporary wetlands exhibit complex hydrology, filling rapidly via local rain storms and then drying through evaporation and groundwater infiltration. Using a long short-term memory (LSTM) neural network to account for these complex processes, we modeled playa inundation for 71,842 playas in the Great Plains from 1984-2018. At the level of individual playas, the model achieved an F1-score of 0.538 on a withheld test set, displaying the ability to predict complex inundation patterns. When averaging over all the playas in the entire region, the model is able to very closely track inundation trends, even during periods of drought. Our results demonstrate potential for using LSTMs to model complex hydrological dynamics. Our modeling approach could be used to model playa inundation into the future under different climate scenarios to better understand how wetland habitats and groundwater will be impacted by changing climate.

**Plain Language Summary**

Playas are small, rain-fed lakes typically found in the Great Plains of the US. Most of the time, they are dry, but when filled they provide important wetland habitat for migrating birds. As the water drains from the playas, they help to recharge the High Plains aquifer, which provides much of the water for agriculture in the region. We used a machine learning model to predict when individual playas are wet and when they are dry using weather, playa size, and information about the land use adjacent to each playa. Our model can accurately predict when playas fill and drain, valuable information for conservation efforts in the region. This research can be used by conservation managers and land-owners to help protect these critical wetlands.

## 1 Introduction

Playas are shallow, depressional wetlands in the western Great Plains that are critical wildlife habitats and sources of groundwater recharge (Smith, 2003). They are essential migratory stopover and wintering habitat for wetland dependent birds in the Central Flyway (Bolen et al., 1989; Davis & Smith, 1998). These temporary wetlands are the lowest point of their own watershed and are filled by intense local precipitation from convective storms. Water is lost through a combination of evapotranspiration and infiltration into the High Plains aquifer (Gurdak & Roe, 2010), a critical and rapidly depleting source of groundwater for agricultural irrigation (Haacker et al., 2016; Steward et al., 2013).

As a result of their semi-arid ecosystem, playas have extended wet-dry cycles, on average only becoming wet once every 11 years (Johnson et al., 2011), but this average masks the wide annual variation in inundation status throughout the region. Furthermore, climate futures for the western Great Plains are warmer, with potential changes in precipitation and humidity that could alter playa hydrology (Ojima et al., 2015; Ojima & Lackett, 2002). Understanding the spatial and temporal patterns of playa inundation—and the ability to predict these patterns into the future— is essential for effective, long-lasting playa conservation. For example, understanding historical patterns of wetness and projecting to the future will allow for targeted conservation to regions that are important now due to their current inundation patterns and regions that will become more important as climate continues to change. Predicting these changes would enable better climate adaptation and risk evaluation for playa-associated wildlife populations.

A variety of previous efforts have focused on predicting inundation as a function of weather, land cover, and other factors (Bartuszevige et al., 2012; Cariveau et al., 2011; Johnson et al., 2011). For example, Bartuszevige et al. (2012) used a generalized linear mixed model (GLMM) to predict inundation and found positive associations with 14-day precipitation, precipitation variance, playa size, and the slope of surrounding terrain. However, these studies are limited spatially and temporally (both within and among years) and thus extrapolation to the entire playa region or across years is problematic.

Advances in analyzing remotely sensed imagery have allowed scientists to investigate patterns of inundation over longer time periods and broader spatial scales. Research has leveraged Landsat 5 satellite observations to investigate how land cover and land cover change relates to inundation, uncovering longer hydroperiods in urban settings and shorter hydroperiods in croplands, rangelands, and grasslands (Collins et al., 2014; Starr & McIntyre, 2020). Common to many of these efforts is the notion that if we can predict historical inundation status as a function of temperature, precipitation, and spatial context, we may be able to predict future inundation and allow for more effective playa conservation.

Two recent innovations motivate our approach for modeling playa inundation. First, the availability of a long term (1984-2019) monthly water history dataset—the JRC Monthly Water History v1.2, hereafter JRC data—provides a novel source of inundation data across the entire playa region (Pekel et al. 2016). Second, the rise of deep learning over the past decade provides new tools for leveraging weather and spatial context to predict playa inundation. In particular, long short term memory (LSTM) neural networks (Hochreiter & Schmidhuber, 1997), have gained traction in a variety of sequence modeling tasks including natural language processing and time series modeling. Time series applications of LSTMs for hydrology include rainfall-runoff modeling (Kratzert et al., 2018; Li et al., 2020) and lake temperature profile modeling (Daw et al., 2020). LSTMs account for both long and short range time dependence, and nonlinear relationships between inputs and outputs. Because these are neural networks, we can also integrate categorical feature embeddings to mop up residual variation and improve predictive power for observed units (Guo & Berkhahn, 2016).

Here we present a monthly inundation model across the playa region that builds upon previous inundation work to evaluate how well we can predict inundation as a function of climate, land cover, and playa-specific features. The LSTM model predicts inundation status derived from the intersection of JRC monthly water history with known playas as a function of high resolution historical climate data and landcover. The resulting model provides high quality monthly predictions of inundation at the playa level while also capturing broad scale regional patterns in inundation. This provides a key step towards an approach that predicts future inundation status and facilitates more effective playa conservation.

## 2 Methods

### 2.1: Data Collection and Preprocessing

The Playa Lakes Joint Venture (PLJV) probable playa data includes 71,842 playa polygons along with key attributes: size, estimated frequency of inundation, distance to the nearest road in feet, and binary flags for if they have been hydrologically modified, if they have been farmed around or through, if they are healthy (playas are considered "healthy" if they are not modified and not farmed around or through), and if they belong to a cluster of playas (PLJV,

2019). The layer also includes the name of the author organization for each playa (who added it to the database).

We calculated playa inundation for every month of the 30m JRC Global Surface Water product from March, 1984 through December, 2018 using Google Earth Engine (Gorelick et al., 2017; Pekel et al., 2016). Inundation was defined as containing one or more pixels that had water detections. The JRC product is highly accurate, with less than 5% errors of omission. However, its performance is worse for small water bodies, seasonal water, and/or water with standing vegetation (Pekel et al., 2016). Since many playas meet one or more of these conditions (e.g. 33% of the PLJV playas are smaller than 0.5 ha), the JRC product likely does not capture all inundations. Nevertheless, the data is generally very accurate and is methodologically consistent throughout the time series, and so we decided to proceed with the JRC data. However, our modeling methods are flexible enough to accommodate any calculated inundation time series. If better and/or higher resolution surface water data becomes available in the future it would be easy to retrain the model using the updated data.

Monthly precipitation, temperature, and vapor pressure deficit (VPD) were obtained from the PRISM Gridded Climate historical data products (PRISM Climate Group, 2020). In the same way we calculated inundation, we extracted monthly weather data for each of the 71,842 playas using Google Earth Engine from 1984 through the end of 2018. This resulted in a time series of 3 weather variables that we could match to the inundation data.

Previous research has shown that the land-use/land-cover (LULC) around a playa can dramatically impact its inundation dynamics (Bartuszevige et al., 2012). To incorporate this into our model, we used modeled and historical USGS LULC data (Sohl et al., 2016, 2014), which is available as a backcasted annual data product from 1938 to 1992 (Sohl et al., 2018b), historical data from 1992 to 2005, and future projections for four different emissions scenarios from the IPCC's Special Report on Emissions Scenarios (the A1B, A2, B1, and B2 scenarios) from 2005 to 2100 (Sohl et al., 2018a). For modeling inundation between 2005 and 2018 we used projections from the A1B scenario, which represents a focus on economic growth with a mixture of energy production in a globalized world (Nakicenovic et al., 2000). Since there is very minimal difference between the SRES scenarios between 2005 and 2018 and the overwhelming majority of divergence begins after 2020, scenario selection was not critical for our 1984-2018 time frame. After downloading the data, we extracted the fraction of each land cover type per year within a 200m buffer of the center point for each playa. We used a random point sampling approach to perform the buffered extraction, generating 5,000 random points within each 200m buffer and then calculating the fraction of those 5,000 that fell within the different LULC classes. This method is efficient and provides accurate estimates of the fraction of the buffer area that falls in each raster pixel, whereas rasterization approaches typically only include or exclude pixels in their entirety based on whether they intersect or majority intersect with the buffer. The code for the randomized buffer extraction is publicly available on GitHub (Solvik, 2020a).

In order to capture watershed information, we used the USGS Watershed Boundary Dataset Hydrologic Unit Codes (HUCs). The units are organized into nested levels of detail, such that the largest units (regions, or HUC-2 because the code contains two digits) are broken down into successive smaller units: subregions (HUC-4), accounting units (HUC-6), and cataloging units (HUC-8, also known as watersheds). The most recent data further subdivides cataloging units into HUC-10 and HUC-12 codes. We downloaded the Watershed Boundary Dataset HUC shapefiles from the USGS website (USDA-NRCS et al., 2020) and performed a spatial join in

QGIS to assign each playa to its HUC-8. In total, the playas fell into 140 different HUC-8 watersheds.

2.2: Modeling

For each month in our data, each playa is either inundated with some amount of water (represented as a 1) or not inundated (0). While this binary approach sacrifices the finer detail of fractional inundation measures, it is more robust to errors in the playa polygons and/or surface water data. It also simplifies the modeling problem.

The data was split into train, validation, and test sets by year: all observations from 1984 through 2010 (27 years) were placed in the training set, observations from 2011 through 2014 (4 years) were placed in validation, and everything from 2015 through 2018 (4 years) was withheld as the final test set. After splitting the data into train, validation, and test sets, we scaled the numerical inputs to the model (e.g. monthly precipitation) by centering the mean to 0 and scaling to unit variance. The scaler was fit to the train set and then applied to the validation and test sets to avoid any information leakage that would unfairly benefit the model.

We included three categorical variables in the model: playa ID (from 0 to 71,842), HUC-8 code (one of 140 options), and playa author (one of 5 authors). Each of these categorical variables were included via embedding layers of dimension 16, 8, and 4, respectively. Briefly, an embedding of dimension D is a matrix that provides a vector of length D for each category. This is a multivariate generalization of "dummy"/indicator variables, which are embeddings of dimension 1. For example, the embedding of playa ID is a matrix with 71,842 rows and D=16 columns. For any one playa, the length 16 vector corresponding to one row can represent features unique to that particular playa. In total, the set of all such vectors can represent features of every particular playa, and account for differences that aren't explained by other model inputs.

We used pytorch's LSTM implementation with binary cross entropy as the loss function. After experimenting with different network sizes and evaluating on the validation set, we settled on a hidden layer with 128 dimensions, an ID embedding of 16 dimensions, HUC-8 embedding of 8 dimensions, and author embedding of 4 dimensions. Excluding the embeddings, the input consisted of 26 features (Table 1). The model was trained using the Adam optimizer with a starting learning rate of 0.01 and a multiplicative decay (gamma) of 0.9 every 5 epochs. Weight regularization was used, with an L2 penalty of $2.5 \times 10^{-6}$. When validation loss did not improve for 16 consecutive epochs, training was halted and the model state from the best-performing epoch was saved. In this case, the model trained for 51 epochs and the model weights at epoch 35 were saved. The LSTM was trained on an AWS EC2 instance with a single NVIDIA T4 GPU.

The data processing and modeling code is publicly available on GitHub (Solvik, 2020b). The input data is publicly available on figshare (Solvik et al., 2020).

**Table 1**. *Input features for LSTM*

| Feature Category | Feature Names |
|---|---|
| Monthly Weather Data (3 features) | Temperature, Precipitation, Vapor Pressure Deficit (VPD) |
| PLJV Playa Attributes (9 features) | Playa area (in acres), mean wet frequency, distance to road (in feet), saturated thickness of the High Plains aquifer in 2013, and binary flags for whether the playa was: healthy, farmed, hydrologically modified, and/or part of a cluster |
| Land Cover Fraction (13 features) | Fraction of each Sohl et al. LULC code: water (1), urban (2), clearcut (3), mining (6), barren (7), deciduous (8), evergreen (9), mixed (10), grassland (11), shrubland (12), cropland (13), pasture (14), wetland (15, 16) |
| Categorical (3 features, passed through embedding layers) | Playa ID (one of 71,842 values), HUC-8 code (one of 140 values), Author (one of 5 values) |

## 3 Results

### 3.1 Playa Inundation Overview

Of the 71,842 playas in the PLJV probable playas dataset, 73% were never inundated during the 34-year time series generated from JRC surface water data (Figure 1). Playas in the southern half of the region (New Mexico, Oklahoma, and Texas) were much more likely to be inundated at least once compared to playas in the north (Colorado, Kansas, and Nebraska). This difference may be explained by the north tending to have smaller (median playa size is 0.78 ha vs. 1.09 ha in the whole dataset) and shallower playas located in croplands (74% farmed vs 63% in the whole dataset). The JRC data has higher omission errors for seasonal water and small water bodies and so it likely overlooks inundation of many of the smaller, ephemeral playas. Out of the playas smaller than the median size, 94% of them were never inundated according to the JRC surface water data. Additionally, farming around the playas dramatically alters their inundation dynamics. Out of the farmed playas, 84% were never inundated, compared to 57% of the unfarmed playas.

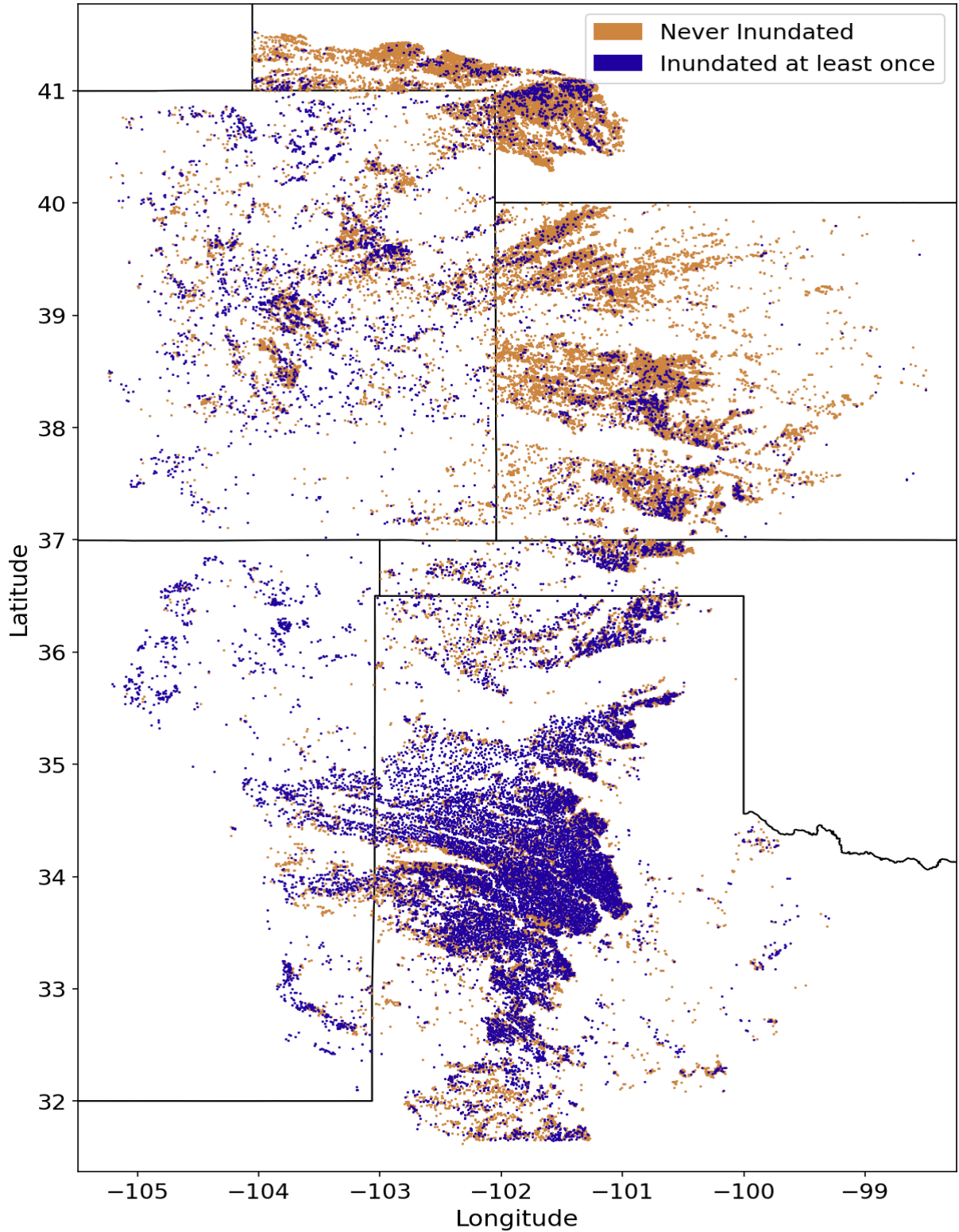

**Figure 1.** Map of the 71,842 playas in the PLJV dataset. Based on the JRC water data, 73% of the playas were never inundated (e.g., at least one pixel with a recorded wetness signature) during our 34-year time series.

3.2: Overall Model Performance/Accuracy

The LSTM was able to achieve good performance on the validation and test sets, with F1-scores of 0.490 and 0.538 respectively. Various performance metrics are shown in Table 2. As expected, the model achieved a higher AUC score and better (i.e. lower) loss on the validation set compared to the test, and the Receiving Operating Characteristic curve (ROC curve, Figure 2) shows excellent performance. However, the test set's precision, recall, and F1-scores were all higher than the validation set. This could be due to the drought period during the validation set, or a result of the cutoff used to decide whether a prediction was a 1 or 0 (we used 0.3 based on validation F1 performance).

**Table 2.** Final model performance on validation and test sets. For calculating precision, recall, and F1, a binary cutoff of 0.3 was used.

|  | **Accuracy** | **BCE Loss** | **AUC** | **Precision** | **Recall** | **F1-Score** |
|---|---|---|---|---|---|---|
| **Validation** | 0.980 | 0.0478 | 0.973 | 0.478 | 0.503 | 0.490 |
| **Test** | 0.961 | 0.090 | 0.962 | 0.487 | 0.601 | 0.538 |

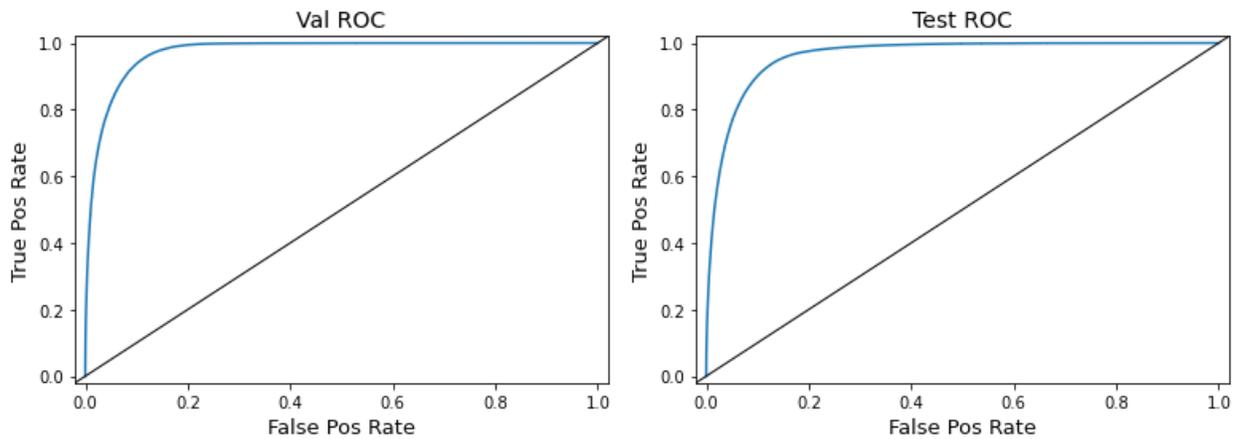

**Figure 2.** Validation and test set receiver operating characteristic (ROC) curve

When we look at model performance for individual playas, performance varies. Generally, performance is excellent for playas that are never inundated during the 34-year time span, a sign that the playa ID embedding is working well. For other playas, the model generally can predict inundation fairly well, particularly for playas with strong seasonal patterns. Figure 3 shows the true and predicted inundation time series for 3 playas: the one with the best test loss, the one with the worst, and the one with the median. In the case of the best performing playa, we see that the model ably learns the strong seasonal pattern that continues through the validation and test time periods. In the case of the worst, the playa is never inundated during the training period and so the model is unable to replicate the sudden inundation pattern that begins in 2014. In the case of the median (and many others), the model identifies most inundation spikes but has a handful of false positives and false negatives throughout.

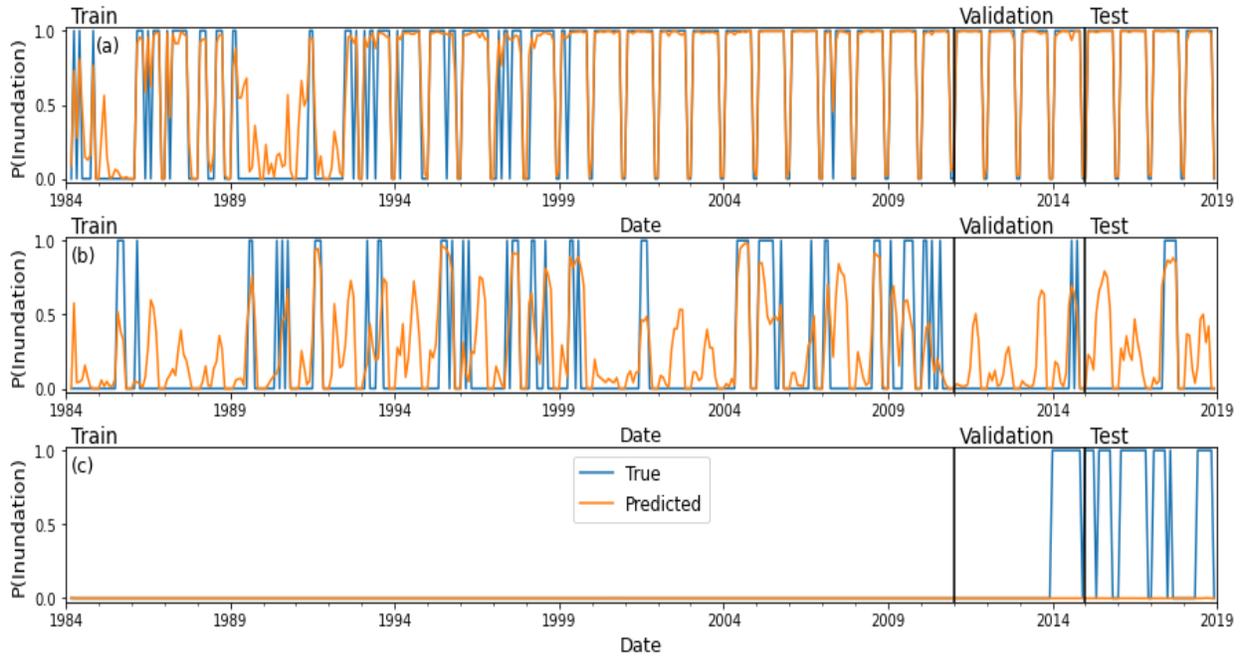

**Figure 3.** Examples of individual playa inundation vs. predicted inundation probability timelines. (a) is the playa with the best test loss (excluding non-inundated playas), (b) is the playa with the median test loss, and (c) is the playa with the worst.

Although individual playa performance varies somewhat, the model performs excellently at the regional scale. Figure 4 shows the predicted inundation fraction (the fraction of playas inundated during any given month) for all playas in the dataset compared to the ground-truth inundation fraction. The predictions track the ground-truth very closely, even during the drought in the validation set (from 2011-2013, at its most extreme in 2012) when inundation drops steeply. Performance decreases a bit during the validation and test periods, but still the predictions match the ground-truth fairly well, a sign that the model is able to successfully generalize to unseen data. Based on these results, it seems that prediction errors for individual playas average out at the regional scale, producing accurate regional inundation estimations even though the model may over or underestimate inundation for individual playas.

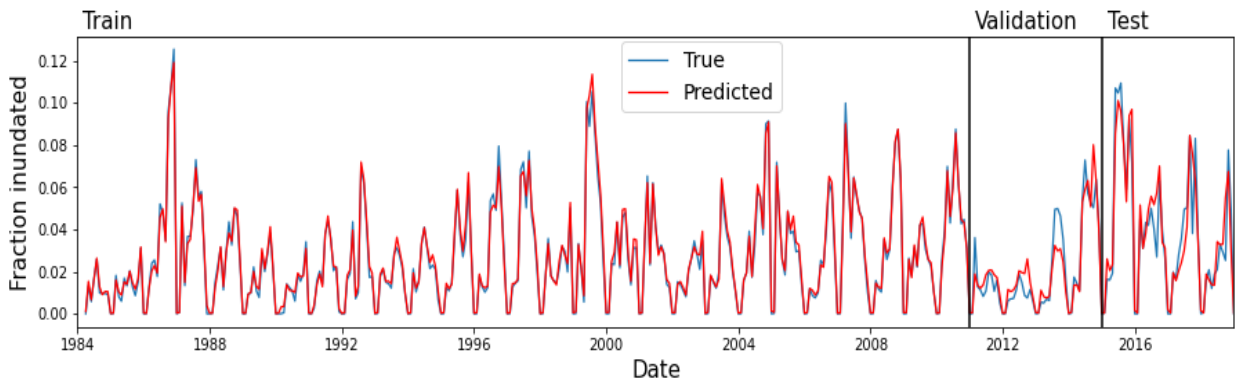

**Figure 4.** Simulated fraction of playas inundated for each month in the time series.

3.3: Spatial Accuracy

We evaluated the test set loss and F1-score for each playa individually. The results are shown in Figure 5. In the case of non-inundated playas, an F1-score cannot be calculated and so those were omitted from the map.

In terms of F1, model performance generally is better at the southern end of the range (New Mexico and Texas). This could be due to the higher percentage of non-inundated playas in the north. Based on the HUC and author embeddings, the model likely learned to generally predict less inundation in the north, and so performance is worse for the relatively small percentage of playas in the region that exhibit at least some inundation.

When looking at the loss, that trend seems to reverse: the model performs worse on the playas in Texas and New Mexico than in Nebraska, Kansas, and Colorado. This is also explained by the high-concentration of non-inundated playas in the north. The model generally handles non-inundated playas very well, and can more easily learn to predict no inundation throughout their entire time series which leads to a low (i.e. better) loss. Recall that non-inundated playas are not included in the F1 map, since F1 is not defined if there are no true positives.

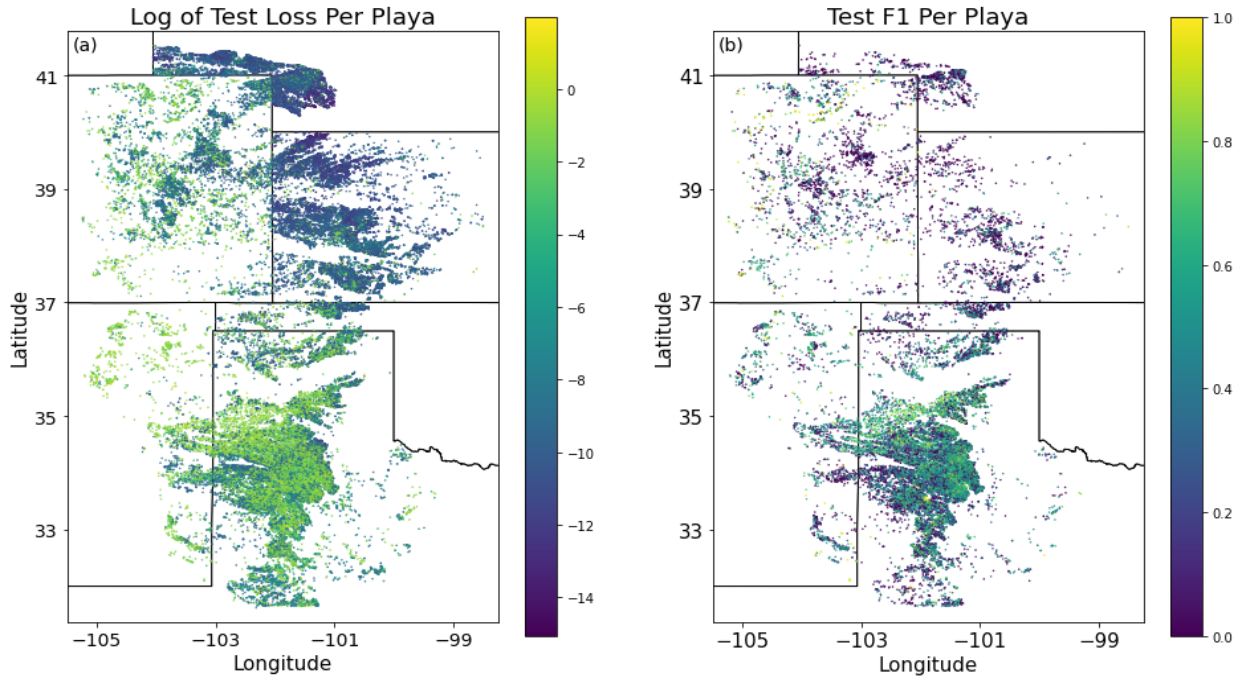

**Figure 5.** Map of test loss (a) and test F1 scores (b) for each playa.

**4 Discussion**

Overall, the LSTM captures inundation dynamics of individual playas fairly well and captures inundation at regional scales even better. Because LSTM models are able to capture complex (e.g., nonlinear and nonstationary) dynamics, this approach lends flexibility to hydrological predictions. If sufficient inundation, meteorological, and landscape data are available, this approach could allow for near term inundation forecasting and playa network time series reconstruction.

We were limited by the resolution and accuracy of available playa, weather, and surface water data. The JRC surface water product has 30m spatial resolution, which is too large to capture precise playa inundation. Further, due to inconsistent Landsat data availability and its 30m resolution, the JRC product performs best on large and permanent water bodies that are clear of vegetation (Pekel et al., 2016). Visual inspection of the JRC data showed that it may frequently miss minor inundation events, particularly for smaller playas, playas which are very rarely inundated, and/or those with standing vegetation. This may largely explain why 73% of the playas were never inundated during our 34-year time series. The PLJV is currently developing a surface water classification approach specifically for detecting playa wetness. If and when that or another higher resolution surface water data product becomes available, it would be easy to retrain our playa model on new inundation data. This would likely produce more accurate inundation predictions, especially for smaller, more ephemeral playas.

The relatively coarse resolution of the JRC data was one reason why we decided to model inundation as a binary variable: no inundation versus at least partial inundation. We originally planned to model inundation as a continuous fraction between 0 (no inundation) and 1 (fully inundated), but the LSTM frequently diverged during training when attempting to model the continuous fraction. Modeling binary inundation proved to be much more stable, and at 30m resolution the JRC surface water was unlikely to capture small changes in inundation fraction anyways. However, binary inundation predictions are less detailed than continuous inundation fraction predictions, especially for large playas where even a 10% difference in inundated area could indicate a sizable difference in water volume. Nevertheless, our binary approach provides playa conversation planners with the most essential information: whether any part of the playa is likely to be wet during a given month.

The LULC and climate data were also limited by their spatial resolution (500m for the Sohl et al. LULC data and 4km for the PRISM climate product). In the case of the LULC data, the land cover immediately surrounding a playa has important effects on inundation (Bartuszevige et al., 2012). With a 500m resolution, the Sohl et al. data may not fully capture variations in LULC near playas. The PRISM climate data does not capture localized rain storms, which may contribute to playa inundation. Further, many playas could fall within the same 4km PRISM grid cell. Higher resolution LULC and climate data would likely boost model performance. Additionally, the LSTM can only model inundation of playas identified in the PLJV probable playas dataset and if there are unmapped playas our approach cannot capture their inundation patterns until they are added to the dataset.

Although the LSTM exhibited good performance, there are limitations associated with black box deep learning methods. First, neural networks in general and LSTMs in particular are not as easy to interpret as more traditional time series models. For example, LSTMs do not readily permit interpretable coefficients that represent the effect of precipitation, temperature, etc. on the model's predictions. Second, this model accounts for unmodeled heterogeneity among playas using playa ID embeddings: learned vector representations that allow the model behavior to vary from playa to playa. These embeddings are learned using training data, and in a time series setting the training data consist of historical observations. In a prediction setting, if the hydrology of a playa abruptly changes due to modification in the future, embeddings may not capture subsequent inundation dynamics because they're learned using historical data. Finally, no hydrological dynamics are included in the model. A physics-guided approach that constrains the LSTM with a science-based dynamical model of inundation might allow for a compromise between the flexibility of a neural network, and the physical consistency of a science-based hydrological model (e.g. Karpatne et al., 2018).

This work points to a number of future directions that could inform habitat conservation in the playas region. Using future projections of climate and land cover data, this model could be used to predict inundation futures at the individual playa and regional scales. These inundation futures are important for understanding how warmer climate futures and/or land cover change might alter inundation, groundwater recharge, availability of critical playa wetland habitat for migratory birds in the Central Flyway (Gitz & Brauer, 2016; Sohl, 2014; Sohl et al., 2012; Starr & McIntyre, 2020). Beyond availability, connectivity is also important for these critical wetlands, and inundation futures could be used to understand the range of spatiotemporal network dynamics (McIntyre & Strauss, 2013).

## 5 Conclusions

Playas are critical wetland habitats in the Great Plains, and predicting inundation as a function of land cover, playa characteristics, and climate is important for understanding regional hydrology and wildlife habitat availability. Using an LSTM and a monthly historical record over three decades, we demonstrate that data-driven prediction of future inundation status is possible, and may exhibit particularly good performance at a regional scale, with a spectrum of results for individual playas.

**Acknowledgments and Data**

The authors thank Earth Lab at University of Colorado Boulder for technical and computing support and the staff of the North Central Climate Adaptation Science Center for their feedback on the research. This research was supported by funding from the US Geological Survey North Central Climate Adaptation Science Center (NC CASC). This work was also made possible by the University of Colorado Boulder Grand Challenge initiative and the Cooperative Institute for Research in the Environmental Sciences through their investment in Earth Lab. The authors do not have any financial conflicts of interest. The dataset for this research is available on figshare (Solvik et al., 2020) under the Creative Commons BY 4.0 license: https://doi.org/10.6084/m9.figshare.13017650.v1.